\documentclass{svproc}
\usepackage{url}
\usepackage{graphicx}
\usepackage{fancyhdr}
\usepackage{multirow} 
\usepackage{multicol} 
\usepackage{tabularx}
\usepackage{adjustbox}
\usepackage{marvosym}

\usepackage[pagebackref=true,breaklinks=true,letterpaper=true,colorlinks,bookmarks=false]{hyperref}
\usepackage{wrapfig}
\usepackage{cite}
\usepackage{amsmath,amssymb,amsfonts}
\usepackage{algorithmic}
\usepackage{graphicx}
\usepackage{textcomp}
\usepackage{tabularx}
\usepackage{url}
\usepackage{times}
\usepackage{epsfig}
\usepackage{graphicx}
\usepackage{amsmath}
\usepackage{amssymb}
\usepackage{stfloats}
\usepackage{multirow}
\usepackage{colortbl}
\usepackage{float}
\usepackage{color}
\usepackage[section]{placeins}

\usepackage{float}
 \usepackage{booktabs}
 \usepackage{multirow}
\usepackage{graphicx}
\usepackage[table,xcdraw]{xcolor}
\begin{document}
\mainmatter              % start of a contribution
\title{Deep Features for Contactless Fingerprint Presentation Attack Detection: Can They Be Generalized?}
\titlerunning{Hamiltonian Mechanics}  % abbreviated title (for running head)
\author{Hailin Li \and Raghavendra Ramachandra\
}
\authorrunning{Hailin \& Raghavendra} % abbreviated author list (for running head)
%Craig Chambers, Kim B. Bruce, and Elisa Bertino}
%

\institute{Norwegian University of Science and Technology (NTNU), Gj\o{}vik, Norway,\\
\email{Email: hailin.li;raghavendra.ramachandra@ntnu.no}%\\ WWW home page:
%\texttt{http://users/\homedir iekeland/web/welcome.html}
%\and
%Universit\'{e} de Paris-Sud,
%Laboratoire d'Analyse Num\'{e}rique, B\^{a}timent 425,\\
%F-91405 Orsay Cedex, France
}

\maketitle              % typeset the title of the contribution

\begin{abstract}
The rapid evolution of high-end smartphones with advanced high-resolution cameras has resulted in contactless capture of fingerprint biometrics that are more reliable and suitable for verification. Similar to other biometric systems, contactless fingerprint-verification systems are vulnerable to presentation attacks. In this paper, we present a comparative study on the generalizability of seven different pre-trained Convolutional Neural Networks (CNN) and a Vision Transformer (ViT) to reliably detect presentation attacks.  Extensive experiments were carried out on publicly available smartphone-based presentation attack datasets using four different Presentation Attack Instruments (PAI). The detection performance of the eighth deep feature technique was evaluated using the leave-one-out protocol to benchmark the generalization performance for unseen PAI. The obtained results indicated the best generalization performance with the ResNet50 CNN. 
\end{abstract}
%%%%%%%%%%%%%%%%%%%%%
\section{Introduction}
%.Fingerprint recognition system has been widely deployed in various applications for many years. Recently, with the rapid revolution of the smartphone camera, contactless fingerphoto verification algorithms achieve promising performance. Attributed to the user convenience and and high acceptance, using smartphone based fingerphoto recognition systems has become a trend that plenty of researchers contributed to this field. Especially under the COVID-19 pandemic, contactless based authentication gained significant attention as an alternative over traditional contact based fingerprint recognition systems. 

%However, fingerphoto recognition system suffers from the presentation attack that the attacker may maliciously spoof the system by producing a fingerprint replica or simply replay/printout the fingerphoto to present it to the camera. An example of different PAIs is demonstrated in Figure \ref{Fig:Pipeline}. After realizing the risks posed by fingerphoto presentation attack, many researchers make efforts and contributions on fingerphoto PAD methods. Table \ref{tab:Smartphone FPAD} summarizes the fingerphoto PAD techniques developed based on smartphone data. In this work, we present our experiments of different deep features for detecting presentation attacks using the latest fingerphoto PAD dataset released by Liveness Detection Competition 2023 For Noncontact Fingerprint Systems and Algorithms \cite{purnapatra2023presentation}. 

Contactless fingerprint verification is widely accepted and deployed in various access control applications. The outbreak of COVID-19 further accelerated the demand for contactless capture devices for the inflection free and reliable person verification. The contactless fingerprint can be captured using dedicated sensors, in which fingerprint images are captured in a contactless fashion and processed further to perform verification. A contactless fingerprint can also be captured using a smartphone, either using a frontal or back camera. The back camera from the smartphone is prominently used to capture fingerprints by considering the high image quality, usability, and good resolution that are useful for extracting minutiae features. The contactless fingerprint captured using smartphones is also referred to as a fingerphoto because the captured fingerprint images are postprocessed using build image processing tools before they are used for biometric authentication. Fingerphoto recognition has been studied extensively and has achieved promising results \cite{ramachandra2023finger}.

\begin{table*}[htbp]
\centering
\resizebox{0.95\columnwidth}{!}{%
\begin{tabular}{|c|c|p{10.355em}|p{12.355em}|c|}
\hline
 Author& Year & Method & Database and PAIs & Generalisation study\\
\hline
 Taneja et al. \cite{taneja2016fingerphoto} & 2016 & Hand-crafted based approach & 1536 bona fide and 4096 spoofed images with  two PAIs & No\\
 \hline
 Zhang et.al \cite{zhang20162d} & 2016 & CNN and hand-crafted based approach & 67011 bona fide and 65581 attack samples with three PAIs& No\\
\hline
Fujio et.al \cite{fujio2018face} &2018 & AlexNet  & 4096 genuine and 8192 spoofed images with one PAI& No\\
\hline
Wasnik et.al \cite{wasnik2018presentation} & 2018 &  LBP, BSIF and HOG with SVM & 50 subjects
consisting of three sessions of bona fide data and three PAIs & No\\
\hline
Marasco and Vurity \cite{marasco2021fingerphoto}  & 2021 & AlexNet, ResNet18 & 4096 genuine and 8192 spoofed images with three PAIs & Yes\\
\hline
Marasco et.al \cite{marasco2022deep}& 2022 & AlexNet, DenseNet201, DenseNet121, ResNet18, ResNet34, MobileNet-V2  & 4096 genuine and 8192 spoofed images with three PAIs& No \\
\hline
Purnapatra et. al \cite{purnapatra2023presentation}& 2023 & DenseNet 121 and NASNet &  14000 bonafide and 1000 attack samples with five PAIs &Yes \\
\hline
\textbf{This Work} & \textbf{2023} & \textbf{AlexNet, DenseNet201, MobileNet-V2, NASNet, ResNet50, GoogleNet, EfficientNet-B0  and Vision Transformers} &  \textbf{5886 bonafide and 4247 attack samples with four PAIs} &\textbf{Yes} \\
\hline
\end{tabular}
}
\label{tab:Smartphone FPAD}
\caption{Existing smartphone based Contactless Fingerprint PAD methods}
\end{table*}

The popularity of smartphone-based fingerprint verification systems has increased their deployment and attracted the attention of attackers. This resulted in the vulnerability of contactless fingerprint or fingerphoto-based biometric systems for different types of Presentation Attack Instruments (PAI), such as 2D and 3D print artifacts, replay attacks, and generating replicas of the fingerprint using different materials. The vulnerability of contactless fingerprint systems has encouraged researchers to develop Presentation Attack Detection (PAD) techniques to improve system reliability. Table \ref{tab:Smartphone FPAD} summarizes the existing techniques for contactless fingerprint PAD, which can be broadly classified as hand-crafted features (e.g., micro-texture, gradients, and light reflection ) and deep features, based on which the pre-trained CNNs are used  for feature extraction. Despite several existing  PAD techniques, the generalization of these techniques for unseen PAI has not been well exploited, particularly with deep features. Readers can refer to a comprehensive survey on the fingerprint PAD \cite{li2023deep} for a more detailed description of the existing PAD. 
Owing to the small size of Presentation Attack (PA) datasets, the use of a pre-trained CNN as a feature extraction method is widely employed in the literature. Therefore, in this work, we present a comprehensive study on the generalizability of pretrained CNNs for unseen PAI. The main contributions of this paper are as follows: (a) We present the deep features extracted using Vision Transformers (ViT) for the fingerprint PAD, (b)  Benchmark eight different pre-trained CNNs on the publicly available fingerprint PA dataset, and (c) Present an extensive analysis with a leave-one-out evaluation protocol to evaluate the generalizability of deep features to unseen PAI.

The rest of the paper is organized as follows: Section 2 describes the proposed scheme for detection task. Section 3 discusses the evaluation protocol and the obtained result. Lastly, we conclude this work in Section 4. 

\begin{figure}[htb!]
\centering
    \includegraphics[width=\textwidth]{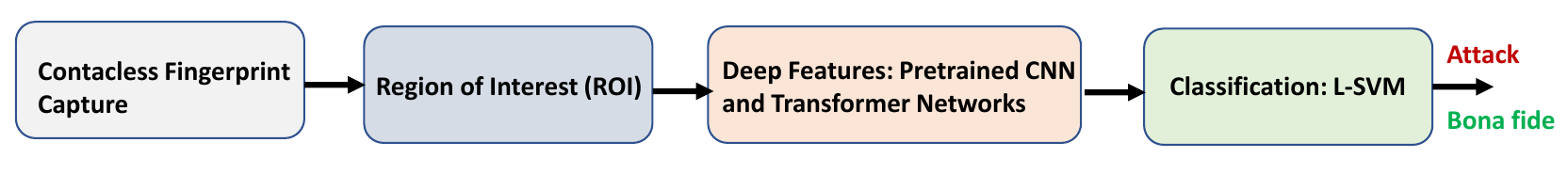}
    \vspace{-0.2in}
    \caption{Block diagram of the deep features based contactless fingerprint PAD}
    \label{Fig:Pipeline}
\end{figure}
\vspace{-0.1in}

%%%%%%%%%%%%%%%%%%%%%%%%%%%
\section{Deep features for contactless fingerprint attack detection}

\label{sec:deepfeatures}
Figure \ref{Fig:Pipeline} shows the block diagram of the deep features based framework for contactless fingerprint verification. Given a fingerprint image, the first step is to extract the Region of Interest (ROI) to discard the background and capture the rich information of the fingerprint. The ROI was captured by approximately locating the center of the finger and cropping the region of dimension $128 \times 256$ pixels. In the next step, the ROI is used to extract the deep features using eight different pre-trained deep learning networks that are trained on the ImageNet dataset that are described below: 
\begin{itemize}  
    \item \textbf{AlexNet \cite{AlexNet}}: is a milestone introduced in 2012 in making deep learning more widely applicable. It employs an 8-layer convolutional neural network consisting of five convolutional layers and three fully-connected layers. In our experiment, we take the feature map after the first fully-connected layer(fc6) and obtain a feature vector of size 4096.
    \item \textbf{GoogLeNet \cite{googleNet}}: introduces the inception module that runs multiple operations (pooling, convolution) with multiple filter sizes in parallel to deal with the over parameterization problem. GoogLeNet has nine inception modules. We utilize the network to obtain the feature vector from the global average pooling layer. 
    \item \textbf{ResNet50 \cite{ResNet}}: is an innovative neural network introduced in 2015 that brings the concept of “skip connections" and residual blocks to tackle the gradient vanishing problem. We utilize the ResNet 50 architecture to extract features from the global average pooling layer.
    \item \textbf{DenseNet201 \cite{DenseNet}}: is a network that connects each layer to every other layer in a feed-forward fashion with fewer parameters and high accuracy compared to ResNet. Thus, the network can strengthen the feature propagation, encourage feature reuse, and substantially reduce the number of parameters. We utilize DenseNet201 to extract features from the global average pooling layer.
    \item \textbf{MobileNetV2 \cite{howard2017mobilenets}}: is a lightweight network containing two main components: Inverted Residual Block and Bottleneck Residual Block and we obtain a size of 1280 feature vector from the global average pooling layer.
    \item \textbf{EffiecientNetb0 \cite{efficientNet}}: introduces a compound scaling method to scale different dimensions of the models to improve performance. We utilize the b0 network of the EffiecientNet network and obtain the feature vector from the global average pooling layer. 
    \item \textbf{NasNet \cite{nasNet}}: frames the problem of finding the best CNN architecture as a Reinforcement Learning problem. The model searches for the best combination of parameters of the given search space. We utilize the pretrained NasNet weight using the matlab deep learning toolbox. 
    \item \textbf{Vision Transformer(ViT) \cite{ViT}}:  split the input image into a sequence of fixed-size non-overlapping patches. Subsequently, a CLS token is added to serve as a representation of the entire image. We utilize the base-sized ViT architecture with a patch resolution of 32 $\times$ 32 pre trained in ImageNet.

\end{itemize} 
In the last step, a linear Support Machine Vector (SVM) is used as the comparator to make the final decision on the given deep features corresponding to the test fingerprint. 
%%%%%%%%%%%%%%%%%%%%%%%%%%%
\section{Experiments and Results}
\label{sec:exp}
\begin{figure}[htb!]
\centering
   \includegraphics[width=0.9\textwidth]{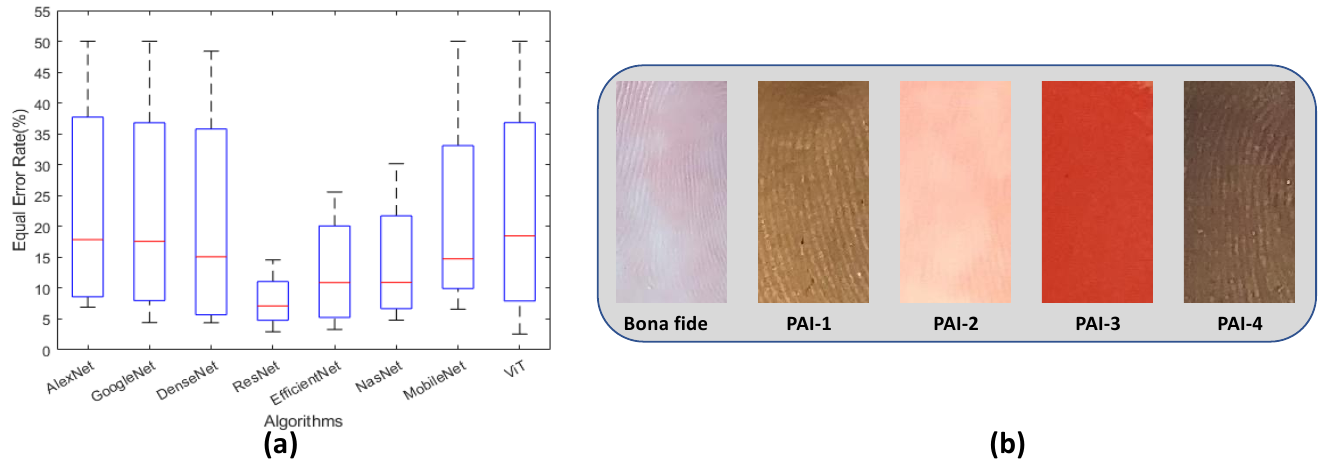}
   \vspace{-0.2in}
   \caption{(a) Box plot distribution indicating the average detection performance (b) example of bonafide and PAIs}
    \label{Fig:Box}
\end{figure}

In this section, we present the quantitative results for eight different deep features for a generalized contactless fingerprint PAD. The experiments are presented using publicly available dataset \cite{purnapatra2023presentation} with four different PAIs (see Figure\ref{Fig:Box} b) such as  ecoflex, playdoh, photo paper and woodglue with 1248, 1623, 1623 and 272 samples respectively that are captured using iPhone X and Samsung Galaxy S20. To benchmark the generalizability of the deep features, we followed the leave-one-out protocol, in which three PAIs were used for training and one PAI was used for testing. Because we have four PAIs, this will result in four cases:  in \textbf{Case-1:}  photo paper, playdoh and woodglue are used for training, ecoflex is used for testing \textbf{Case-2:} ecoflex, playdoh, and woodglue are used for training, photo paper is used for testing\textbf{ Case-3:} ecoflex, photo paper and woodglue are used for training, playdoh is used for testing \textbf{Case-4:} ecoflex, photo paper, and playdoh are used for training, and woodglue is used for testing.

\begin{table*}[htp]

\caption{Quantitative performance of the deep features for contactless fingerprint PAD}
\vspace{-0.1in}
\label{tab:PADtable}
\resizebox{1\columnwidth}{!}{%
\begin{tabular}{|
>{\columncolor[HTML]{EFEFEF}}l |l|l|ll|
>{\columncolor[HTML]{DAE8FC}}l |l|l|ll|}
\hline
\cellcolor[HTML]{EFEFEF}                                   &                                  &                         & \multicolumn{2}{l|}{BPCER @ APCER =} & \cellcolor[HTML]{DAE8FC}                                  &                                  &                         & \multicolumn{2}{l|}{BPCER @ APCER =} \\ \cline{4-5} \cline{9-10} 
\cellcolor[HTML]{EFEFEF}                                   & \multirow{-2}{*}{PAD Algorithms} & \multirow{-2}{*}{D-EER} & \multicolumn{1}{l|}{5\%}    & 10\%   & \cellcolor[HTML]{DAE8FC}                                  & \multirow{-2}{*}{PAD Algorithms} & \multirow{-2}{*}{D-EER} & \multicolumn{1}{l|}{5\%}      & 10 \%    \\ \cline{2-5} \cline{7-10} 
\cellcolor[HTML]{EFEFEF}                                   & AlexNet                          & 6.91                    & \multicolumn{1}{l|}{11.67}  & 3.78   & \cellcolor[HTML]{DAE8FC}                                  & AlexNet                          & 25.43                   & \multicolumn{1}{l|}{61.94}  & 47.98  \\ \cline{2-5} \cline{7-10} 
\cellcolor[HTML]{EFEFEF}                                   & GoogleNet                        & 11.54                   & \multicolumn{1}{l|}{22.24}  & 13.47  & \cellcolor[HTML]{DAE8FC}                                  & GoogleNet                        & 23.63                   & \multicolumn{1}{l|}{59.28}  & 43.86  \\ \cline{2-5} \cline{7-10} 
\cellcolor[HTML]{EFEFEF}                                   & DenseNet201                      & 6.97                    & \multicolumn{1}{l|}{10.22}  & 4.43   & \cellcolor[HTML]{DAE8FC}                                  & DenseNet201                      & 23.19                   & \multicolumn{1}{l|}{63.18}  & 50.20  \\ \cline{2-5} \cline{7-10} 
\cellcolor[HTML]{EFEFEF}                                   & ResNet50                         & 6.65                    & \multicolumn{1}{l|}{8.28}   & 3.50   & \cellcolor[HTML]{DAE8FC}                                  & ResNet50                         & 14.58                   & \multicolumn{1}{l|}{30.22}  & 20.85  \\ \cline{2-5} \cline{7-10} 
\cellcolor[HTML]{EFEFEF}                                   & EfficientNet-B0                  & 7.21                    & \multicolumn{1}{l|}{11.15}  & 4.47   & \cellcolor[HTML]{DAE8FC}                                  & EfficientNet-B0                  & 14.58                   & \multicolumn{1}{l|}{34.05}  & 21.47  \\ \cline{2-5} \cline{7-10} 
\cellcolor[HTML]{EFEFEF}                                   & NasNet                           & 13.30                   & \multicolumn{1}{l|}{29.62}  & 17.73  & \cellcolor[HTML]{DAE8FC}                                  & NasNet                           & 30.16                   & \multicolumn{1}{l|}{83.59}  & 69.01  \\ \cline{2-5} \cline{7-10} 
\cellcolor[HTML]{EFEFEF}                                   & MobileNet-V2                     & 12.00                   & \multicolumn{1}{l|}{29.10}  & 15.14  & \cellcolor[HTML]{DAE8FC}                                  & MobileNet-V2                     & 16.22                   & \multicolumn{1}{l|}{49.40}  & 28.00  \\ \cline{2-5} \cline{7-10} 
\multirow{-10}{*}{\cellcolor[HTML]{EFEFEF}\textbf{Case-1}} & ViT              & 6.71                    & \multicolumn{1}{l|}{8.17}   & 3.74   & \multirow{-10}{*}{\cellcolor[HTML]{DAE8FC}Case-2}         & ViT 
               & 50.00                   & \multicolumn{1}{l|}{100.00} & 100.00 \\ \hline \hline \hline
\cellcolor[HTML]{FFCE93}                                   & AlexNet                          & 50.00                   & \multicolumn{1}{l|}{97.03}  & 93.99  & \cellcolor[HTML]{FFCCC9}                                  & AlexNet                          & 10.30                   & \multicolumn{1}{l|}{18.84}  & 10.31  \\ \cline{2-5} \cline{7-10} 
\cellcolor[HTML]{FFCE93}                                   & GoogleNet                        & 50.00                   & \multicolumn{1}{l|}{96.93}  & 93.01  & \cellcolor[HTML]{FFCCC9}                                  & GoogleNet                        & 4.43                    & \multicolumn{1}{l|}{4.29}   & 1.88   \\ \cline{2-5} \cline{7-10} 
\cellcolor[HTML]{FFCE93}                                   & DenseNet201                      & 48.42                   & \multicolumn{1}{l|}{96.01}  & 91.94  & \cellcolor[HTML]{FFCCC9}                                  & DenseNet201                      & 4.40                    & \multicolumn{1}{l|}{3.98}   & 1.47   \\ \cline{2-5} \cline{7-10} 
\cellcolor[HTML]{FFCE93}                                   & ResNet50                         & 7.58                    & \multicolumn{1}{l|}{11.56}  & 5.49   & \cellcolor[HTML]{FFCCC9}                                  & ResNet50                         & 2.94                    & \multicolumn{1}{l|}{0.26}   & 0.05   \\ \cline{2-5} \cline{7-10} 
\cellcolor[HTML]{FFCE93}                                   & EfficientNet-B0                  & 25.58                   & \multicolumn{1}{l|}{72.78}  & 55.06  & \cellcolor[HTML]{FFCCC9}                                  & EfficientNet-B0                  & 3.30                    & \multicolumn{1}{l|}{1.10}   & 2.51   \\ \cline{2-5} \cline{7-10} 
\cellcolor[HTML]{FFCE93}                                   & NasNet                           & 4.81                    & \multicolumn{1}{l|}{4.66}   & 1.75   & \cellcolor[HTML]{FFCCC9}                                  & NasNet                           & 8.55                    & \multicolumn{1}{l|}{13.24}  & 6.54   \\ \cline{2-5} \cline{7-10} 
\cellcolor[HTML]{FFCE93}                                   & MobileNet-V2                     & 50.00                   & \multicolumn{1}{l|}{98.80}  & 96.51  & \cellcolor[HTML]{FFCCC9}                                  & MobileNet-V2                     & 6.6                     & \multicolumn{1}{l|}{7.85}   & 4.08   \\ \cline{2-5} \cline{7-10} 
\multirow{-8}{*}{\cellcolor[HTML]{FFCE93}\textbf{Case-3}}  & ViT               & 23.65                   & \multicolumn{1}{l|}{100.00} & 100.00 & \multirow{-8}{*}{\cellcolor[HTML]{FFCCC9}\textbf{Case-4}} & ViT                & 2.57                    & \multicolumn{1}{l|}{0.99}   & 0.31   \\ \hline
\end{tabular}
}
\end{table*}

\begin{figure}[htb!]
\centering
   \includegraphics[width=0.9\textwidth]{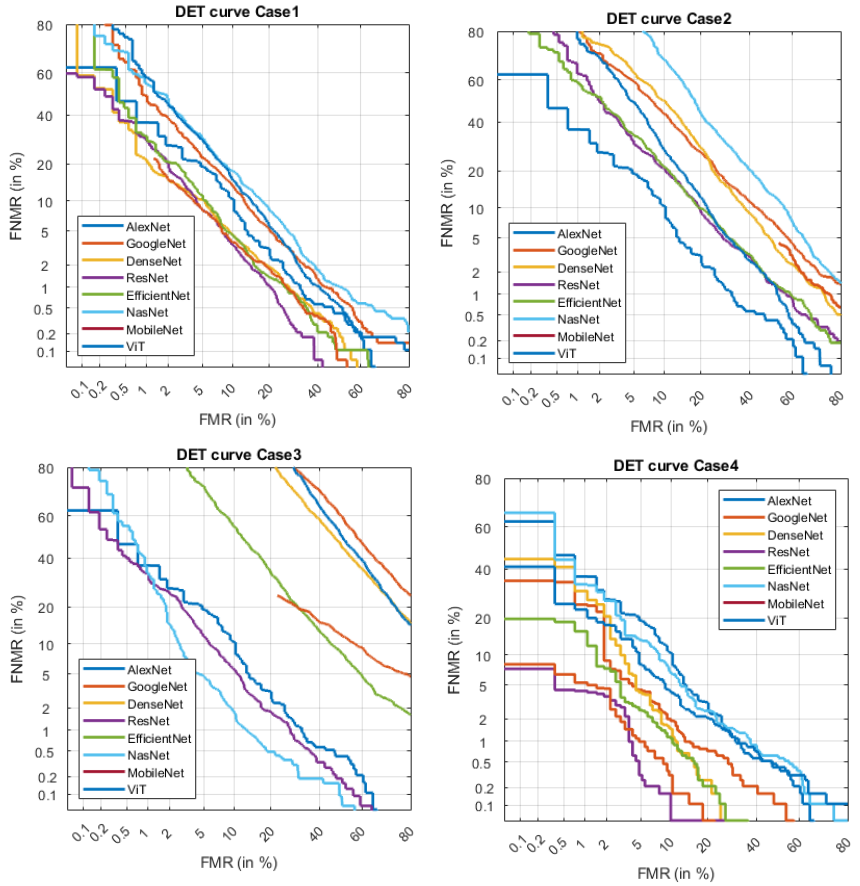}
   \vspace{-0.2in}
   \caption{DET curves of different cases (1 to 4) by comparing different deep learning models}
    \label{Fig:DET}
\end{figure}

Table \ref{tab:PADtable} shows the quantitative performance of eight different pre-trained deep learning networks employed for contactless fingerprint PAD.  The quantitative results are presented using ISO/IEC  30107- 3 \cite{ISO-IEC-30107-3-PAD-metrics-170227} metrics, such as the Attack Presentation Classification Error Rate (APCER) and bona fide Presentation Classification Error Rate (BPCER). Furthermore, D-EER(\%) is also included, which indicates an error when APCER=BPCER. Therefore, the lower the D-EER, the higher the performance. The obtained results highlight the challenge of generalizing deep features to detect unseen PAI. Among the four different cases, Case-1 and 4 indicate the improved performance of all eight deep features towards generalization, as there is a smaller drop in performance between these two cases. Thus, the generalization of  deep features to detect fingerprint PAI generated using  Ecoflex and Woodglue materials is acknowledged  by the reported experiments. The detection performance of the deep features was degraded in both Case-2 and Case-3. Therefore, the unseen detection of photo paper and playdoh is challenging compared to  the Ecoflex and Woodglue materials used to generate contactless fingerprint PAI. It can be noted (see Table \ref{tab:PADtable}) that the deep features extracted  using ViT, AlexNet, GoogleNet, and DenseNet201 indicate severe degradation, thereby indicating the challenge of generalizing to Case-2 and 3.

Figure \ref{Fig:DET} shows a trade-off between the False Match Rate(FMR) and False Non Match Rate(FNMR) which represents the overall performance comparison among different models. Figure \ref{Fig:Box} (a) shows the individual performance (D-EER(\%)) of the deep features averaged over all four cases. A lower D-EER indicates better generalization of the deep feature algorithm towards unseen PAI. Based on the results, ResNet50 exhibits the best generalized detection performance, with the lowest average D-EER of 8.26\%. The highest performance degradation was noted for AlexNet, with an average D-EER of 23.16\%.

%\begin{wrapfigure}{r}{1\textwidth}
 % \begin{center}
  %  \includegraphics[width=0.5\textwidth]{boxPlot.png}
  %\end{center}
  %\vspace{-0.2in}
  %\caption{Box plot distribution among deep features}
%\end{wrapfigure}    boxPlot2

%%%%%%%%%%%%%%%%%%%%%%%%%%%

\section{Conclusion}
\label{sec:conc}

Contactless fingerphoto recognition systems have become popular and are deployed in many smartphone authentication applications. However, it is necessary to increase the system security to be robust against presentation attacks. In this study, we used eight advanced pre-trained deep learning models to evaluate the generalizability against unseen attacks. The latest non-contact fingerphoto PAD dataset was utilized for the experiment using the leave-one-out evaluation protocol. By comparing the obtained APCER, BPCER, and D-EER, we observed that almost all the models were vulnerable to the PAIs of photo paper and playdoh when they were not included in the training data, according to the high EER rate. In contrast, the models have high generalization against Ecoflex and Playdoh. Additionally, ResNet 50 achieves the best result with an average EER of 8.26\%. 
%%%%%%%%%%%%%%%%%%%%%%%%%%%%%%%%%%%

\bibliographystyle{splncs03}
\bibliography{reference}
\end{document}